\documentclass[letterpaper]{article} 
\usepackage{aaai23}

\usepackage{times}  
\usepackage{helvet}  
\usepackage{courier}  
\usepackage[hyphens]{url}  
\usepackage{graphicx} 
\usepackage{natbib}  
\usepackage{caption}
\usepackage{multirow}
\usepackage{amsmath,amssymb}
\usepackage{algorithm}
\usepackage{algorithmic}
\usepackage{url}            
\usepackage{booktabs}       
\usepackage{amsfonts}       
\usepackage{microtype}      
\usepackage{xcolor}         
\usepackage{newfloat}
\usepackage{listings}
\DeclareCaptionStyle{ruled}{labelfont=normalfont,labelsep=colon,strut=off} 
\floatstyle{ruled}
\newfloat{listing}{tb}{lst}{}
\floatname{listing}{Listing}
\title{Model and Data Agreement for Learning with Noisy Labels}
\author{
    Yuhang Zhang,
    Weihong Deng,
    Xingchen Cui,
    Yunfeng Yin,
    Hongzhi Shi,
    Dongchao Wen\textsuperscript{\rm}\thanks{Corresponding author.}
}
\affiliations{

    Inspur Electronic Information Industry Co., Ltd\\
    Shandong Massive Information Technology Research Institute\\
    Jinan, China\\
    zyuhang219@gmail.com, dengweihong2020@foxmail.com, \{cuixingchen, yinyunfeng, shihzh, wendongchao\}@inspur.com
}

\usepackage{bibentry}

\begin{document}

\maketitle

\begin{abstract}
Learning with noisy labels is a vital topic for practical deep learning as models should be robust to noisy open-world datasets in the wild. The state-of-the-art noisy label learning approach JoCoR fails when faced with a large ratio of noisy labels. Moreover, selecting small-loss samples can also cause error accumulation as once the noisy samples are mistakenly selected as small-loss samples, they are more likely to be selected again. In this paper, we try to deal with error accumulation in noisy label learning from both model and data perspectives. We introduce mean point ensemble to utilize a more robust loss function and more information from unselected samples to reduce error accumulation from the model perspective. Furthermore, as the flip images have the same semantic meaning as the original images, we select small-loss samples according to the loss values of flip images instead of the original ones to reduce error accumulation from the data perspective. Extensive experiments on CIFAR-10, CIFAR-100, and large-scale Clothing1M show that our method outperforms state-of-the-art noisy label learning methods with different levels of label noise. Our method can also be seamlessly combined with other noisy label learning methods to further improve their performance and generalize well to other tasks. The code is available in https://github.com/zyh-uaiaaaa/MDA-noisy-label-learning.
\end{abstract}

\section{Introduction}

The performance improvement of Deep Neural Networks (DNNs) depends largely on the large-scale training datasets. However, collecting large-scale datasets with fully precise annotations is usually expensive and time-consuming. There are lots of label noise in the open-world datasets, which degrades the performance of deep learning models in practical applications. At the same time, it is widely known that deep neural networks can easily memorize large-scale data with even completely random labels, making unreliable predictions when generalizing to other tasks~\cite{zhang2021understanding, arpit2017closer, jiang2018mentornet}. Thus, learning with noisy labels has drawn lots of attention in recent years~\cite{patrini2017making, han2018masking, zhang2018generalized, wangyisen2019iccv, junnan2019cvpr, ren2018learning, li2020dividemix, jiang2018mentornet, malach2017decoupling, han2018co, yu2019icml, wei2020combating}. 

Some previous works try to seek theoretically guaranteed methods to solve the noisy label learning problem~\cite{patrini2017making, han2018masking, zhang2018generalized, wangyisen2019iccv}. However, they are usually not suitable for real-world label noise as the transition probability between the noisy label and ground truth in real-world datasets is hard to estimate. Others follow the small-loss selection path~\cite{jiang2018mentornet, malach2017decoupling, han2018co, yu2019icml, wei2020combating}. They treat the small-loss samples as clean samples and only use them for training, while there exists the debate that whether the model should utilize "agreement" or "disagreement". \cite{malach2017decoupling, yu2019icml} claim that only using the samples that two models have different predictions as training samples can keep the models distinct and avoid fitting noisy labels. However, ~\cite{wei2020combating} argues that disagreement is not necessary and proposes a training paradigm that pushes two models together. We observe that~\cite{wei2020combating} achieves excellent performance when learning with a small ratio of noisy labels while it fails with a large ratio of noisy labels. We argue that this might be because the performance of the two models are both low faced with a large ratio of noisy labels, which leads to error accumulation from the model perspective. In the meantime, selecting small-loss samples as clean samples for training leads to error accumulation from the data perspective. Some noisy samples might get small loss values during training and thus be selected for training. After updating the gradients, the selected noisy samples might be remembered by the models and are more likely to be selected again in the following training epochs, which degrades the model performance.

Motivated by the above two problems, in this paper, we propose to deal with the error accumulation in noisy label learning from both model and data perspectives. For the model perspective, we introduce mean point ensemble to utilize a more robust loss function and guide the model to acquire more information from the unselected samples. From the data perspective, we find that the model fits noisy samples through memorization, while it cannot remember the flipped noisy samples to the noisy labels though they are with the same semantic meaning. Thus, we propose to utilize the flipped images to better detect noisy labels for the first time to reduce error accumulation from the data perspective. Though flip is a common augmentation method, using the loss values of flipped images to detect noisy labels is rarely explored. We also show that using flip to detect noisy samples can bring more benefits than simply using flip as a basic augmentation under label noise.
 

We evaluate our proposed method Model and Data Agreement (MDA) on image classification datasets CIFAR-10, CIFAR-100, and large-scale Clothing1M. Extensive experiments validate the effectiveness of each of the two proposed modules. MDA is also compatible with other state-of-the-art noisy label learning methods to further improve their performance and can generalize well to other noisy label learning tasks.

The main contributions of our work are as follows:

\begin{itemize}
\item[1.]We introduce mean point ensemble to utilize a more robust loss function and guide the two models to use information from the unselected samples. We find that the flipped samples can help the model to better detect noisy samples. We also validate that using flip to detect noisy labels achieves better results than simply using flip as a data augmentation method under the task of learning with label noise.
\item[2.]Extensive experiments show that the proposed method advances state-of-the-art noisy label learning methods on noisy label CIFAR-10, CIFAR-100 and large-scale noisy dataset Clothing1M. 
\item[3.]The proposed method can be easily combined with other noisy label learning methods to further improve their performance and generalize well to other noisy label learning tasks.
\end{itemize}

\begin{figure*}[t]
    \centering
        \includegraphics[width=1.0\textwidth]{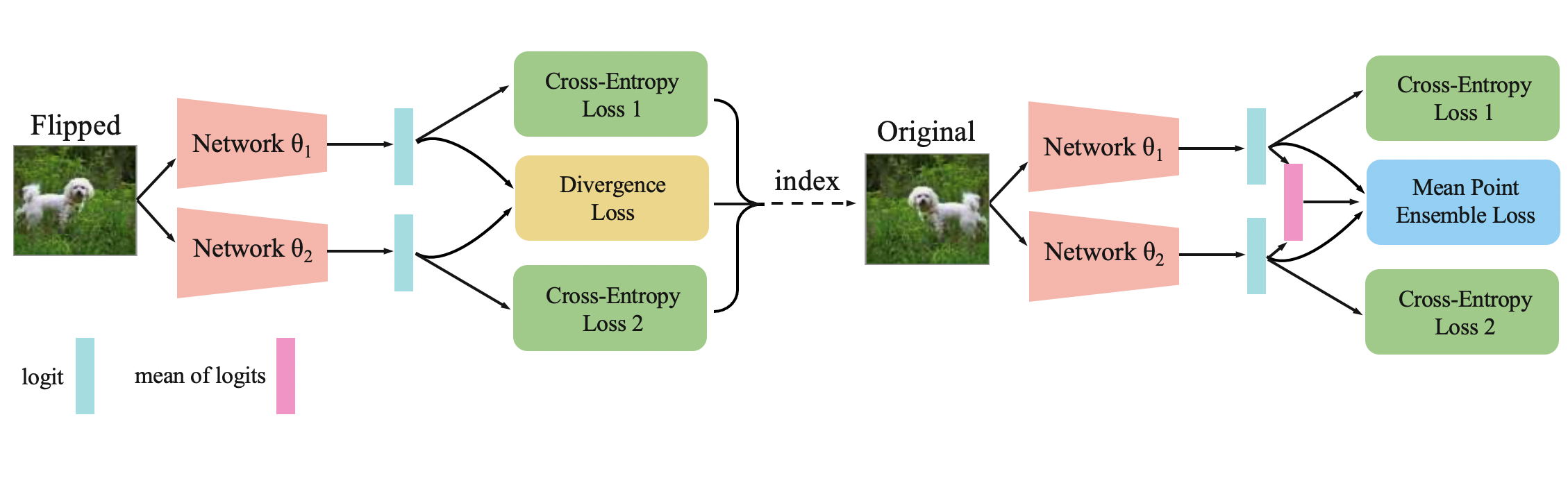}
        \caption{The framework of our MDA. We utilize classification loss and divergence loss to select small-loss samples on the flipped images. We compute classification loss and mean point ensemble loss on the original images to train the two models.}
        \label{fig:mda} 
\end{figure*}

\section{Related Work}
\label{sec:related}

\subsection{Noisy Label Learning}
How to achieve good performance learning with noisy labels has drawn lots of attention in recent years~\cite{patrini2017making, han2018masking, zhang2018generalized, thulasidasan2019combating, xu2019l_dmi, jiang2018mentornet, ren2018learning, arazo2019unsupervised, han2018co, malach2017decoupling, wei2020combating, xie2021partial, yi2019probabilistic, kim2019nlnl, huang2019o2u, han2019deep, li2020dividemix, ye2020purifynet, nguyen2019self, li2021learning}. The noisy label learning methods in recent years can be mainly categorized into two types.

\noindent\textbf{Theory Guaranteed Noisy Label Learning Methods} The first type of noisy label learning method estimates the noise transition matrix to model the label transition probability or propose generalized cross-entropy loss. Patrini \emph{et al.}~\cite{patrini2017making} propose loss correction methods that estimate the label noise transition matrix through training with noisy datasets. Hendrycks \emph{et al.}~\cite{Hen2018nips} estimate the noise transition matrix by using a small set of trusted data. Han \emph{et al.}~\cite{han2018masking} incorporate human cognition of invalid class transitions to help estimate the noise transition matrix. Zhang \emph{et al.} ~\cite{zhang2018generalized} propose noise-robust
loss functions that can be seen as a generalization of mean absolute error (MAE) loss and cross entropy (CCE)
loss, which can utilize both the merits of MAE loss and CCE loss. Xu \emph{et al.}~\cite{xu2019l_dmi} design a novel loss function based on the mutual information theory, which is provably robust to instance-independent label noise. The common merit of these methods is that they are theoretically guaranteed. However, they might be unsuitable for real-world datasets with noisy labels as they usually do not conform to the theoretical assumptions of these methods.

\noindent\textbf{Sample Selection Label Learning Methods}
 The other type of noisy label learning approach is based on the observation that DNNs learn simple patterns before memorizing the noisy labels~\cite{arpit2017closer}. They treat the small-loss samples as clean samples to train the DNNs and filter out large-loss samples as they are likely to be noisy samples. Jiang \emph{et al.}~\cite{jiang2018mentornet} train a mentor net to imitate a human teacher. The mentor net selects clean samples to teach the student network to avoid remembering noisy samples. Han \emph{et al.}~\cite{han2018co} train two differently initialized models to teach each other as the two different models can mitigate different types of errors caused by noisy labels. Malach \emph{et al.}~\cite{malach2017decoupling} update the parameters only on the instances that the two models have different predictions to maintain divergence. Yu \emph{et al.}~\cite{Yu2019arxiv} combine co-teaching~\cite{han2018co} with decoupling~\cite{malach2017decoupling} to further improve the performance of co-teaching as updating only using the samples with different predictions can keep the two models distinct. Wei \emph{et al.} ~\cite{wei2020combating} argue that keeping two models distinct is not necessary and they propose a method named JoCoR to push two models closer during training and use their agreement degree to select small-loss samples. Though the agreement of two models can help to detect noisy samples, we argue that JoCoR fails when facing lots of noisy labels. We propose mean point ensemble to utilize a more robust loss function and further guide the two models to use more information from the unselected images, which stable the training and improves the model performance under a large ratio of label noise.

\subsection{Model Ensemble} 
Model ensemble is a very effective technique that can significantly improve the performance of deep learning models. By ensembling multiple models together, we can reduce the bias and overfitting of a single model, which can make deep learning models more robust to noisy labels. Deep model ensemble methods in supervised learning ~\cite{ganaie2021ensemble} can be mainly categorised into bagging~\cite{breiman1996bagging}, boosting~\cite{zhang2008rotboost}, negative correlation learning~\cite{ncl}, explicit/implicit ensembles~\cite{srivastava2014dropout, wan2013regularization, huang2016deep, singh2016swapout}, homogeneous/heterogeneous ensemble~\cite{breiman2001random, li2018heterogeneous}, decision fusion strategies~\cite{ju2018relative}. Several aforementioned noisy label learning methods can be viewed as utilizing model ensemble methods. Co-teaching~\cite{han2018co} can be viewed as two different initialized models that teach each other, which is a kind of ensemble. JoCoR~\cite{wei2020combating} also uses model ensemble, which is named Deep Mutual Learning~\cite{zhang2018deep}. Deep Mutual Learning makes several student models learn collaboratively and
teach each other throughout the whole training process. JoCoR incorporates Deep Mutual Learning to let the two models teach each other to achieve high performance through model ensemble. However, when facing a large ratio of label noise, the performance of both models is low. Thus, letting the two models teach each other brings the drawback that the two models might learn the errors from each other. We propose mean point ensemble to utilize a more robust loss function and more information from unselected samples to improve the ensemble performance under a large ratio of noisy labels.

\section{Proposed Method}
In this section, we illustrate the implementation details of our proposed Model and Data Agreement (MDA) method. We propose to deal with the noisy label learning problem from both model and data perspectives. Specifically, we introduce mean point ensemble to utilize a more robust loss function and more information from unselected samples. We further utilize the flipped images which are unseen during the training process to select clean samples, which is superior to just use flip as a data augmentation.

\begin{algorithm*}[t]
\caption{Model and Data Agreement}
\label{alg:MDA}
\begin{algorithmic}[1]
\REQUIRE original training set $D$ and the flipped counterpart $\widetilde{D}$, Network $f$ with $\Theta = \{\Theta_1, \Theta_2\}$, learning rate $\eta$, noise rate $\tau$, warm up selection epochs $T_k$ and total epochs $T_{\text{max}}$, data loader iteration $I_{\text{train}}$;
\FOR{$t$ = 1,2,\ldots,$T_{\text{max}}$}
 
    \STATE $\boldsymbol{p}_1$ = $f(\widetilde{\boldsymbol{x}},\Theta_1)$, $\forall \widetilde{\boldsymbol{x}} \in \widetilde{D}$;
    \STATE $\boldsymbol{p}_2$ = $f(\widetilde{\boldsymbol{x}},\Theta_2)$, $\forall \widetilde{\boldsymbol{x}} \in \widetilde{D}$;

    \STATE \textbf{Calculate} the selection loss $\ell_{sel}$ for each sample by \eqref{joint};
        
    \STATE \textbf{Obtain} the indexes of small-loss set on $\widetilde{D}$ according to the ratio $R(t)$;
    
    \FOR{$n = 1,\ldots,I_{\text{train}}$}    
        \STATE \textbf{Fetch} mini-batch $D_n$ from $D$;
        \STATE \textbf{Obtain} $\ell_{cls}$ by \eqref{cls} on $D_n$ samples with indexes fall into the small-loss set and $\ell_{ens}$ by \eqref{ncl} on all the samples;
        
        \STATE \textbf{Obtain} $\ell_{train}$ by \eqref{joint2};
        \STATE \textbf{Update} $\Theta = \Theta - \eta \nabla \ell_{train}$;
    
    \ENDFOR
    \STATE \textbf{Update} $R(t) = 1 - \min \left\{ \frac{t}{T_k} \tau, \tau \right\}$
\ENDFOR
\ENSURE $\Theta_1$ and $\Theta_2$
\end{algorithmic}
\end{algorithm*}

\subsection{Model and Data Agreement}
Given the dataset $D = \{\textbf{x}_i, y_i\}^N_{i=1}$ with $N$ samples, 
we first carry out flip augmentation to all the samples and get the flipped dataset denoted as $\widetilde{D}$. In the training stage, we use $\widetilde{D} = {\{ \widetilde{\textbf{x}}_i, {y}_i \} }^N_{i=1} $ to compute the classification loss and utilize the divergence loss to select clean samples for training, while we only backpropagate gradients on the original dataset $D$. 

Specifically, we also utilize the agreement maximization principles~\cite{icmlworkshop, wei2020combating} to detect noisy samples. We use the joint loss (\ref{joint}) of classification and agreement to select clean samples. 
\begin{equation} l_{sel}(\widetilde{\textbf{x}}_i) =  l_{cls}(\widetilde{\textbf{x}}_i, y_i) + \lambda * l_{ag}(\widetilde{\textbf{x}}_i),\label{joint}\end{equation}
where $l_{cls}(\widetilde{\textbf{x}}_i, y_i)$ means the sum of the classification loss of the two models of sample $\widetilde{\textbf{x}}_i$ and $l_{ag}(\widetilde{\textbf{x}}_i)$ represents the agreement level of the two models towards the prediction results of the sample $\widetilde{\textbf{x}}_i$. They are computed following (\ref{cls}) and (\ref{ag}) respectively. $\lambda$ is the weight of the agreement (divergence) loss when selecting the clean samples.
\begin{equation} 
\begin{split}
    l_{cls} & 
    = l_{C1} + l_{C2},\label{cls}
\end{split}
\end{equation} where

\begin{equation}
l_{C1} = -\frac1N \sum_{i=1}^N  (\log{\frac{e^{f^{1}_{y_i}(\widetilde{\textbf{x}}_i)}}{\sum\nolimits_{j}^M e^{f^{1}_{j}(\widetilde{\textbf{x}}_i)}}}),
\end{equation}

\begin{equation}
  l_{C2} = - \frac1N \sum_{i=1}^N  (\log{\frac{e^{f^{2}_{y_i}(\widetilde{\textbf{x}}_i)}}{\sum\nolimits_{j}^M e^{f^{2}_{j}(\widetilde{\textbf{x}}_i)}}}),
\end{equation}
$f^{1}$ and $f^{2}$ are the outputs of input images $\{\widetilde{\textbf{x}}_i\}_{i=1}^N$ from the two models. $N$ and $M$ represent the number of samples and classes.

\begin{equation} l_{ag} =  D_{KL}(f^{1} || f^{2}) +  D_{KL}(f^{2} || f^{1}),\label{ag}\end{equation}
where \begin{equation}  D_{KL}(f^{1} || f^{2}) = \sum_{i=1}^N \sum_{j=1}^M f_j^{1}(\widetilde{\textbf{x}}_i)\log{\frac{f_j^{1}(\widetilde{\textbf{x}}_i)}{f_j^{2}(\widetilde{\textbf{x}}_i)}} ,\label{kl1}\end{equation}

\begin{equation}  D_{KL}(f^{2} || f^{1}) = \sum_{i=1}^N \sum_{j=1}^M f_j^{2}(\widetilde{\textbf{x}}_i)\log{\frac{f_j^{2}(\widetilde{\textbf{x}}_i)}{f_j^{1}(\widetilde{\textbf{x}}_i)}} .\label{kl1}\end{equation}
When selecting clean samples according to the loss values. Batch size is a very important factor to be considered. Selecting small-loss samples in the mini-batch scale is less effective than selecting them from the whole training set. To reduce the randomness of the sample selection process which might cause performance degradation, we first compute the loss values on the whole dataset $\widetilde{D}$. We then select clean samples according to the selection loss $l_{sel}$ values of all the training samples. We mark the indexes of the small-loss samples and then backpropagate gradients only on the corresponding samples from the original dataset ${D}$. The details of the sample selection are shown in Algorithm~\ref{alg:MDA}.

\subsection{Mean point ensemble}
We introduce mean point ensemble to utilize a more robust loss function and guide the two models to explore more information from unselected samples. The mean point ensemble regularization loss can be simplified to the following 
\begin{equation} l_{ens} = \frac1N \sum_{i=1}^N ((f^{1}(\textbf{x}_i) - \bar{f}(\textbf{x}_i))^2 + (f^{2}(\textbf{x}_i) - \bar{f}(\textbf{x}_i))^2 ) ,\label{ncl}\end{equation}
where \begin{equation}\bar{f}(\textbf{x}_i) = \frac12(f^{1}(\textbf{x}_i) + f^{2}(\textbf{x}_i)).\label{mean}\end{equation}
 The mean prediction results $\bar{f}(\textbf{x}_i)$ can be viewed as the ensemble of two models. Note that we use a symmetric mean squared error loss, which is shown to be more robust to noisy labels ~\cite{ghosh2017robust}. The other difference between our loss and the JoCoR loss is that we calculate $l_{ens}$ on all training samples rather than only the selected samples. Though the unselected samples are more likely to have noisy labels, simply ignoring them might cause information loss. Information loss has been previously validated as an important factor to the model performance of quantization methods\cite{qin2020forward, qin2022bibert, qin2022distribution, qin2020bipointnet}. Mean point ensemble can better explore the information of unselected samples to regularize the consistency of the two models and further improve their performance.

\subsection{Overall Training Loss}
Having acquired the indexes of the clean samples from the flipped images, we use them to index the clean samples from the original training images. We then compute the joint loss of classification on selected samples and the loss of mean point ensemble on all samples to train the two models.
\begin{equation} l_{train} = l_{cls} + \gamma * l_{ens},\label{joint2}\end{equation}
$\gamma$ is the weight of the mean point ensemble loss of the two models, which we will study in the ablation study.

\begin{table*}[t]
\centering
\caption{Average test accuracy (\%) on \textsl{CIFAR-10} over the last 10 epochs. Each experiment is run five times, shown with the mean and standard deviation.}\label{tab:cifar10_data}
\begin{tabular}{c|c|c|c|c|c|c}
\hline
Noise Rate& Baseline &  Decoupling & Co-teaching & Co-teaching+ & JoCoR & MDA\\
\hline
20\% &  $69.18 \pm 0.52$  & $69.32 \pm 0.40$ & $78.23 \pm 0.27$ & $78.71 \pm 0.34$ & $ 85.73 \pm 0.19$ & $ \textcolor{red}{86.08} \pm 0.14$ \\

50\% &  $42.71 \pm 0.42$ &  $40.22 \pm 0.30$ & $71.30 \pm 0.13$ & $57.05 \pm 0.54$ & $ 79.41 \pm 0.25$ & $ \textcolor{red}{80.80} \pm 1.96$ \\

80\% &  $16.24 \pm 0.39$ &  $15.31 \pm 0.43$ & $26.58 \pm 2.22$ & $24.19 \pm 2.74$ & $ 27.78 \pm 3.06$ & $ \textcolor{red}{40.76} \pm 5.41$ \\

\hline

\end{tabular}
\end{table*}

\begin{table*}[t]
\centering
\caption{Average test accuracy (\%) on \textsl{CIFAR-100} over the last 10 epochs. Each experiment is run five times, shown with the mean and standard deviation.}\label{tab:cifar100_data}
\begin{tabular}{c|c|c|c|c|c|c}

\hline

Noise Rate& Baseline &  Decoupling & Co-teaching & Co-teaching+ & JoCoR & MDA\\
\hline
20\% &  $35.14 \pm 0.44$  & $33.10 \pm 0.12$ & $43.73 \pm 0.16$ & $49.27 \pm 0.03$ & $53.01 \pm 0.04$ & $\textcolor{red}{56.44} \pm 0.13$ \\

50\% &  $16.97 \pm 0.40$ &  $15.25 \pm 0.20$ & $34.96 \pm 0.50$ & $40.04 \pm 0.70$ & $43.49 \pm 0.46$ & $\textcolor{red}{47.91} \pm 0.31$\\

80\% &  $4.41 \pm 0.14$  & $3.89 \pm 0.16$ & $15.15 \pm 0.46$ & $13.44 \pm 0.37$ & $15.49\pm 0.98$ & $\textcolor{red}{23.80} \pm 1.19$\\

\hline
\end{tabular}
\end{table*}

\begin{table*}[t]
\centering
\caption{Best test accuracy (\%) and Last epoch test accuracy (\%) on Clothing1M. MDA outperforms JoCoR in both the best and last epoch test accuracy. MDA also gets more consistent best test accuracy and last epoch test accuracy.}\label{tab:clothing}
\begin{tabular}{c|c|c|c|c|c|c}
\hline
Accuracy & Baseline &F-correction&  Decoupling & Co-teaching & JoCoR & MDA\\

\hline
Best &  $67.22$  &$68.93$& $68.48$ & $69.21$  & $70.30$ & $\textcolor{red}{71.24}$ \\
Last &  $64.68$ & $65.36$& $67.32$ & $68.51$ & $69.79$ & $\textcolor{red}{71.07}$\\
\hline
\end{tabular}
\end{table*}

\section{Experiments}
In this section, we illustrate the implementation details of MDA. We then verify MDA on the CIFAR-10, CIFAR-100~\cite{krizhevsky2009learning} and Clothing1M~\cite{xiao2015learning} with different levels of label noise. The ablation studies are carried out to study the effectiveness of the flip noise detection and mean point ensemble separately. We further display some visualization results to provide an intuitive understanding of flip noise detection. We also show that MDA is compatible with other noisy label learning methods and can further improve their performance.

\begin{table*}[t]
\setlength{\tabcolsep}{5pt}
\begin{center}
\caption{Evaluation of the different modules of MDA on CIFAR-100 with different levels of label noise. The results show that both flip noise detection and mean point ensemble can improve the performance from the baseline, while using them together can achieve the best performance.}
\label{table:modules}
\begin{tabular}{ccccc}
\hline
flip noise detection & mean point ensemble & 20\%  & 50\%  & 80\%  \\ \hline
\text{\sffamily x}   & \text{\sffamily x}  & 35.14\%  &     16.97\%    &     4.41\% \\
\checkmark  & \text{\sffamily x}  & 54.84\%   &     45.41\%&  23.73\% \\
\text{\sffamily x}  & \checkmark  & 55.12\%   &     47.39\% &  23.22\% \\

\checkmark  & \checkmark  & \textcolor{red}{56.34\%}   &     \textcolor{red}{48.57\%} & \textcolor{red}{23.97\%}  \\
 \hline
\end{tabular}
\end{center}
\end{table*}

\subsection{Implementation Details}
To make a fair comparison with JoCoR~\cite{wei2020combating}, we use a 7-layer CNN network architecture for CIFAR-10, CIFAR-100~\cite{krizhevsky2009learning} and ResNet-18~\cite{he2016deep} for Clothing1M~\cite{xiao2015learning} as the backbone network. We \emph{do not} use any data augmentation tricks. For CIFAR-10 and CIFAR-100, the batch size is $512$. The initial learning rate is $0.001$. We use Adam~\cite{kingma2014adam} optimizer with the weight decay as $0.0001$. We run $200$ epochs in total and linearly decay the learning rate to zero from $80$ to $200$ epochs. We select small-loss samples following the ratio $R(t)$ as: $R(t) = 1 - min(\frac{t}{T_k}\tau, \tau)$. Following Co-teaching~\cite{han2018co} and JoCoR~\cite{wei2020combating}, we assume that the noise rate $\tau$ is known. $t$ is the current training epoch and we set the $T_k$ as 10 for CIFAR-10 and CIFAR-100 to make a fair comparison with other methods. Following\cite{wei2020combating}, we search the divergence loss weight  $\lambda$ in [0.05, 0.10, 0.15,. . .,0.95] with a clean validation set, $\lambda$ is set to 0.65. We search the weight of the mean point ensemble loss $\gamma$ in [0.001, 0.01, 0.1, 1, 10] with the validation set and finally set $\gamma$ to 1. For experiments on Clothing1M, we train the models for 15 epochs in total. During the training stage, we set the learning rate to $0.0008$ , $0.0005$ and $0.00005$ for $5$ epochs each. We use Adam optimizer (momentum=0.9) and set the batch size to 64 following~\cite{wei2020combating}. Experiments are conducted on 4 NVIDIA RTX 2080Ti GPUs.

\subsection{Evaluation of MDA on CIFAR-10 and CIFAR-100 with Noisy Labels}
We quantitatively evaluate the improvement of our proposed MDA against other state-of-the-art methods. We train our model on the same noisy dataset as other methods to make fair comparisons. We report the mean test accuracy of the last 10 epochs. We follow the tradition to run each experiment 5 times and report the mean and the standard deviation of the accuracy. We explore the robustness of MDA with three levels of label noise including the ratio of 20\%, 50\%, 80\% on CIFAR-10 and CIFAR-100. 

As shown in Table~\ref{tab:cifar10_data}, our method outperforms all other state-of-the-art label noise learning methods by a non-trivial margin. For example, MDA outperforms JoCoR under 20\%, 50\%, 80\% label noise by 0.35\%, 1.39\%, 12.98\% respectively on CIFAR-10. MDA outperforms JoCoR under 20\%, 50\%, 80\% label noise by 3.43\%, 6.47\%, 8.31\% on CIFAR-100. MDA improves state-of-the-art methods more on large noise ratio, which implies that our method can deal with harder noisy datasets. We owe the large improvements on 80\% label noise to the introduction of mean point ensemble and flip noise detection, which reduces error accumulation from both model and data perspectives.

\subsection{Evaluation of MDA on Large-Scale Noisy Data Clothing1M}
To further evaluate the effectiveness of our proposed MDA, we carry out experiments on the large-scale real-world noisy dataset Clothing1M. The experiment results are shown in Table~\ref{tab:clothing}. MDA outperforms JoCoR by 0.94\% and 1.28\% on the best test accuracy and last epoch test accuracy respectively. As Clothing1M is a very large dataset containing 1 million samples, our proposed MDA can be considered as outperforming JoCoR by a non-trivial margin. More importantly, compared with JoCoR, the best test accuracy during the training process and the last epoch test accuracy of MDA is more similar, which means our method can be more robust to the label noise and less likely to overfit the noisy labels for the last several epochs.

\subsection{The Effectiveness of Flip Noise Detection}

\label{flip}

\begin{figure}[h]
    \centering

        \includegraphics[width=0.4\textwidth]{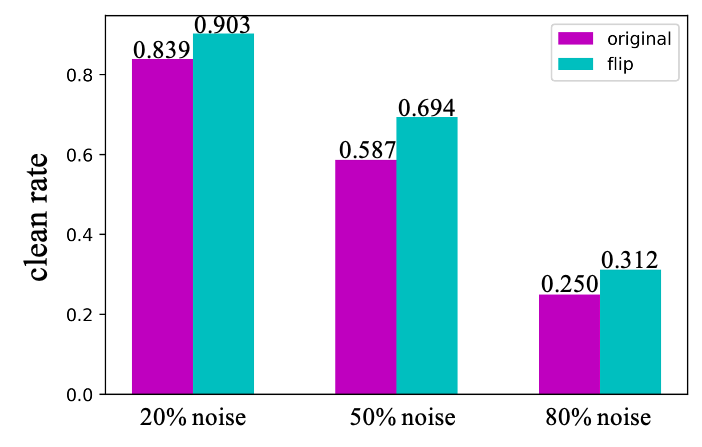}
        \caption{Effectiveness of flip noise detection. We can get cleaner train sets selecting on the flipped images.}
        \label{fig:flip} 
\end{figure} 

In this section, we illustrate the effectiveness of flip noise detection. We train a classification model for 40 epochs using CIFAR-100 with different levels of label noise. We then load the trained model to select small-loss samples using the loss values of original images and flip images separately. The experiment result is shown in Figure~\ref{fig:flip}. When training with 20\% label noise, we select 80\% small-loss samples as clean samples using original images or flip images. The result shows that 83.9\% of the small-loss original samples are without noisy labels while 90.3\% of the small-loss flip images are samples without noisy labels. This experiment illustrates the reason why we select small-loss samples using the flipped ones during training can improve the classification accuracy. We also carry out experiments to compare with the JoCoR method which simply uses flip augmentation for noisy label training. Under 80\% label noise in CIFAR-100, using flip augmentation only gets 19.40\% accuracy, while using flip to detect noise gets 23.80\%, which illustrates that simply using flip augmentation has relatively small improvement on training with noisy labels.

\begin{figure*}[t]
\centering
\includegraphics[width=1.0\textwidth]{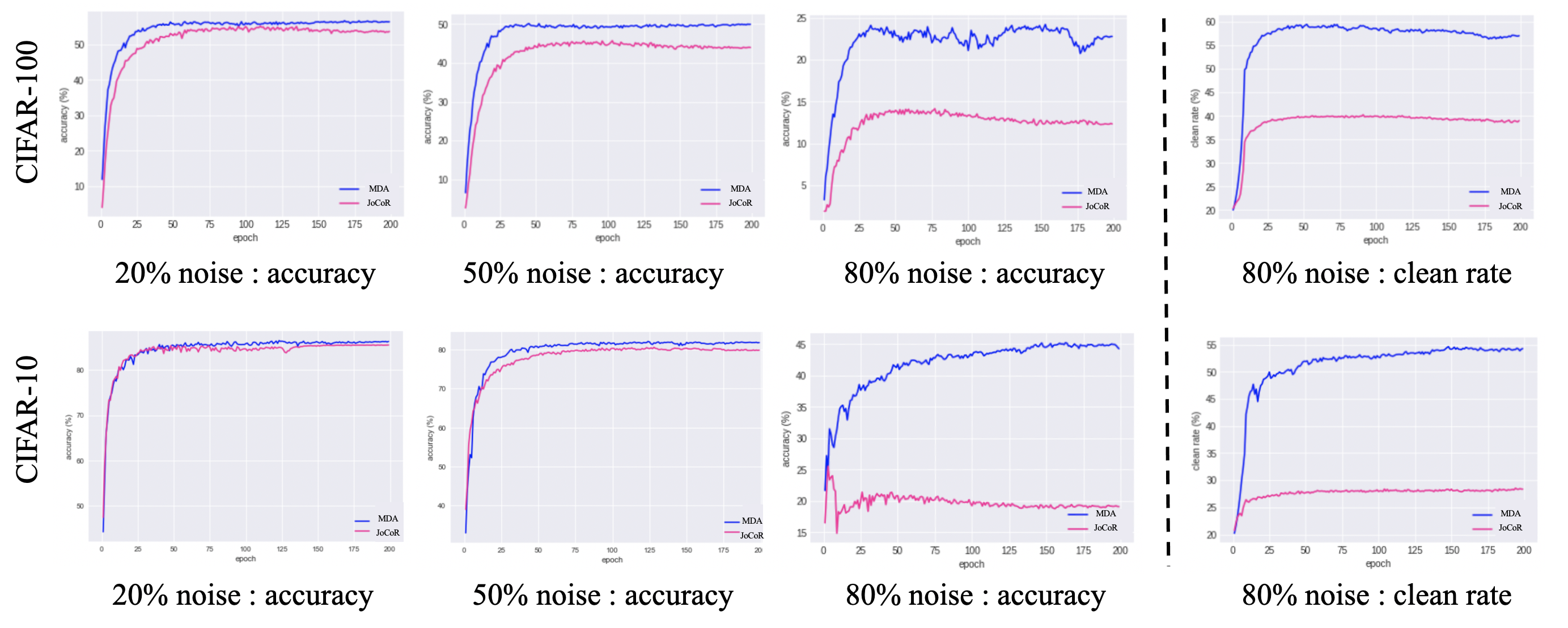}
\caption{The test accuracy of JoCoR and MDA on CIFAR-10 and CIFAR-100. MDA outperforms JoCoR under all different noise levels. We also plot the clean rate of the selected samples of the two methods learning with 80\% label noise. MDA can select a cleaner training set because of the flip noise detection and mean point ensemble.}
\label{fig:values}
\end{figure*}
\begin{table*}[t]
\begin{center}
\setlength{\tabcolsep}{1.8pt}
\caption{Combine with Co-teaching. We seamlessly combine the flip noise detection and mean point ensemble modules with the existing noisy label learning method Co-teaching and observe that the two modules can further improve the performance of Co-teaching.}
\label{table:generalization}
\begin{tabular}{c|cccccc}
\hline
\multirow{2}{*}{Methods} & \multicolumn{3}{c|}{CIFAR10}                                 & \multicolumn{3}{c}{CIFAR100}                             \\  \cline{2-7} 
                         & \multicolumn{1}{c}{20\% noise} & \multicolumn{1}{c}{50\% noise} & \multicolumn{1}{c}{80\% noise} & \multicolumn{1}{c}{20\% noise} & \multicolumn{1}{c}{50\% noise} & \multicolumn{1}{c}{80\% noise} \\ \hline
Baseline                 &    69.38\%                 &       42.70\%               &      16.55\%                 &    35.68\%                 &        17.38\%               &         4.24\%              \\
Co-teaching                    &     77.61\%                   &    71.16\%                    &         33.21\%               &               44.10\%         &        35.15\%                &    14.81\%                  \\
Co-teaching + MDA                     &       \textcolor{red}{84.76\%}                 &       \textcolor{red}{73.22\%}                 &       \textcolor{red}{33.24\%}                 &    \textcolor{red}{55.48\%}                    &      \textcolor{red}{43.91\%}                  &          \textcolor{red}{26.32\%}              \\ \hline
\end{tabular}
\end{center}
\end{table*}

\begin{table*}[!h]
\setlength{\tabcolsep}{6pt}
\begin{center}
\caption{MDA on noisy label FER tasks. MDA can also deal with other noisy label learning tasks. We compare MDA with Co-teaching and other state-of-the-art FER noisy label learning methods.}
\label{table:sota}
\begin{tabular}{lccc}
\hline
Method      & 10\% label noise & 20\% label noise & 30\% label noise \\ \hline
Baseline    &    81.01\%        &  77.98\%          &  75.50\%          \\ 
SCN    &    82.15\%        &     79.79\%       &     77.45\%       \\
RUL         &  86.17\%          &    84.32\%        &      82.06\%      \\
EAC         &    \textcolor{red}{88.02\%}        &     \textcolor{red}{86.05\%}       &      \textcolor{red}{84.42\%}      \\ \hline
Co-teaching &   83.57\%         &    81.75\%        &    79.78\%        \\
MDA (Ours)  &     \textcolor{red}{87.77\%}        &     \textcolor{red}{87.18\%}         &   \textcolor{red}{84.57\%}         \\ \hline
\end{tabular}
\end{center}
\end{table*}

\subsection{Evaluation of Different Modules}
\label{abl}
To show the influence of the flip noise detection module and the mean point ensemble module separately, we carry out an ablation study on CIFAR-100 with different levels of label noise. The results are shown in Table~\ref{table:modules}. Without the flip noise detection and mean point ensemble modules, the model degrades to the baseline method. When we detect the label noise through the flipped images, the test accuracy on the test set increases compared with the baseline, which illustrates that using the unseen flipped images can help the model to better filter out noisy samples. When we only use mean point ensemble, the robust loss function and more information from unselected samples improve the model performance. The results also show that using flip noise detection or mean point ensemble alone can both help the model to achieve higher performance than baseline while using the two modules together achieves the best performance. We conclude that the two modules can be used separately to deal with noisy labels while they can achieve the best performance when they cooperate.

\subsection{Visualization of Accuracy and Clean rate}
We plot the test accuracy versus training epochs under different noisy labels in Figure~\ref{fig:values}. When the noise rate is 20\% or 50\%, JoCoR can achieve high performance and the improvement from our proposed method is limited. However, when it comes to 80\% noise, we can clearly view that JoCoR fails in both CIFAR-10 and CIFAR-100 and achieves very low performance. The results show that our method outperforms JoCoR by a large margin when the noise rate is large. We speculate the reason lies in that the flip noise detection module selects more clean samples from data perspective and the mean point ensemble utilize a more robust loss function and more information from unselected samples to improve the model performance. We further plot the clean rate of JoCoR and our method under a large noise rate, it aligns with the test accuracy curves, which confirms that our method selects more clean samples.

\subsection{Combine with Other Noisy Label Learning Methods}
MDA can also be combined with other state-of-the-art noisy label learning methods to further improve their performance. As MDA deals with noisy labels from model and data perspectives, we can improve other small-loss selection methods with MDA. Specifically, when selecting noisy samples, we could use the loss values of the flipped images instead of the original training images to reduce the error accumulation of sample selection from the data perspective. When training two models, we can combine the mean point ensemble with other methods to ensemble the models utilizing a more robust loss function and the information of unselected samples. Co-teaching maintains two different initialized models which select clean samples for each other to suppress the error accumulation. We add our proposed method to Co-teaching to improve its performance to show the plug-in and play characteristics of our proposed method. We select clean samples using flipped images and add the mean point ensemble loss during the whole training process of Co-teaching. The results are shown in Table~\ref{table:generalization}. Without bells and whistles, Co-teaching plus MDA achieves the best performance under all different noise levels on both CIFAR-10 and CIFAR-100 datasets, which illustrates that MDA can be seamlessly combined with sample selection noisy label learning methods to further improve their performance. 

\section{The Generalization Ability of MDA}
MDA can also be utilized to solve other tasks. Facial Expression Recognition (FER) aims at helping computers to understand human behavior or even interact with a human by recognizing human expression. As we use flip noise detection to filter out the noisy samples, our method can generalize to other tasks with images of the same semantic meaning before and after the flip. 
The images of facial expression recognition are also symmetric. Thus, we carry out experiments on the FER dataset RAF-DB~\cite{li2017reliable} and compare our method with several state-of-the-art FER noisy label learning methods. 

The results are shown in Table~\ref{table:sota}, we compare MDA with Co-teaching~\cite{han2018co}. We also display some state-of-the-art FER noisy label learning methods~\cite{zhang2021relative, zhang2022learn, wang2020suppressing}, they do not need the exact noise rate to filter out the exact ratio of large-loss samples like Co-teaching and MDA. The results imply that our proposed MDA can acquire state-of-the-art performance knowing the exact noise rate compared with other FER noisy label learning methods. Furthermore, Co-teaching also needs to know the exact noise rate to filter out large-loss samples, while it does not generalize well to FER noisy label learning task as its performance is outperformed by MDA by a large margin under all the different noise levels.

\section{Conclusion}

In this paper, we deal with noisy label learning from both model and data perspectives. We introduce mean point ensemble to utilize a more robust loss function and guide the two models with the information from unselected samples. We also find that the flipped images can be utilized to better detect noisy samples and achieves better performance than just using flip as an augmentation method. Extensive experiments validate that our proposed MDA method outperforms other state-of-the-art noisy label learning methods and each of the modules improves the JoCoR method. Furthermore, MDA can be seamlessly combined with other noisy label learning methods to further improve their performance and generalize well to other noisy label learning tasks.

\bibliography{anonymous-submission-latex-2023.bbl}

\end{document}